\documentclass{article}
\usepackage[margin=.9in]{geometry}

\usepackage[utf8]{inputenc} % allow utf-8 input
\usepackage[T1]{fontenc}    % use 8-bit T1 fonts
\usepackage{hyperref}       % hyperlinks
\usepackage{url}            % simple URL typesetting
\usepackage{booktabs}       % professional-quality tables
\usepackage{amsfonts}       % blackboard math symbols
\usepackage{nicefrac}       % compact symbols for 1/2, etc.
\usepackage{microtype}      % microtypography
\usepackage{xcolor}         % colors

\usepackage{graphicx} % Required for inserting images
\usepackage{subfig}
\usepackage{amsmath}
\usepackage{physics}
\usepackage{upgreek}
\usepackage{times}
\usepackage[backend=biber,style=numeric-comp,sorting=none]{biblatex} % Ensures citations appear in order and compress
\usepackage{authblk}
\usepackage[capitalize,nameinlink]{cleveref}

\title{Learning with springs and sticks}

\author[1,2]{Luis Mantilla Calderón\thanks{Email: luis@cs.toronto.edu}}
\author[1,2,3,4,5,6,7]{Alán Aspuru-Guzik}
\affil[1]{\footnotesize Department of Computer Science. University of Toronto. 40 St George St., Toronto, ON M5S 2E4, Canada}
\affil[2]{\footnotesize Vector Institute for Artificial Intelligence. W1140-108 College St., Schwartz Reisman Innovation Campus, Toronto, ON M5G 0C6, Canada}
\affil[3]{\footnotesize Department of Chemistry. University of Toronto.  80 St. George St., Toronto, ON M5S 3H6, Canada}
\affil[4]{\footnotesize Department of Chemical Engineering \& Applied Chemistry. University of Toronto. 200 College St., Toronto, ON M5S 3E5, Canada}
\affil[5]{\footnotesize Department of Materials Science \& Engineering, University of Toronto. 184 College St., Toronto, ON M5S 3E4, Canada}
\affil[6]{\footnotesize Acceleration Consortium. 700 University Ave., Toronto, ON M7A 2S4, Canada}
\affil[7]{\footnotesize NVIDIA. 431 King St. W \#6th, Toronto, ON M5V 1K4, Canada}

\begin{document}

\maketitle

\begin{abstract}
    Learning is a physical process. Here, we aim to study a simple dynamical system composed of springs and sticks capable of arbitrarily approximating any continuous function. The main idea of our work is to use the sticks to mimic a piecewise-linear approximation of the given function, use the potential energy of springs to encode a desired mean squared error loss function, and converge to a minimum-energy configuration via dissipation. We apply the proposed simulation system to regression tasks and show that its performance is comparable to that of multi-layer perceptrons. In addition, we study the thermodynamic properties of the system and find a relation between the free energy change of the system and its ability to learn an underlying data distribution. We empirically find a \emph{thermodynamic learning barrier} for the system caused by the fluctuations of the environment, whereby the system cannot learn if its change in free energy hits such a barrier. We believe this simple model can help us better understand learning systems from a physical point of view. 
    % We study the system's behavior at different scales and find a \emph{thermodynamic learning barrier} that depends on the environment for which the system cannot learn if crossed. Moreover, we do simulations to study the approximation error of the model, calculate its entropy production throughout training, and solve regression on a reduced MNIST dataset. We believe this simple model can help us better understand learning systems from a physical point of view. 
\end{abstract}

\section{Introduction}

% Current artificial intelligence systems have shown us that the ability to learn an underlying probability distribution from samples up to a desired fidelity is a property of a physical system. GPUs, TPUs, and other machine-learning-tailored hardware, are used as universal simulators for intelligent systems. However, due to the rapid development of this new field, there has been relatively short time to explore hardware and software optimizations. This has led to severely energy-inefficient procedures for training large models. A clear example is the vast amount of CO$_2$ emissions for training a model like BERT \cite{strubell2019energy}, which logically has been largely amplified with larger models such as GPT-4, Gemini, StableDiffusion XL, and others. The comparison between current energy requirements of these models to those of living beings evidence a challenge that remains unsolved, and nonetheless, will require novel engineering.
% Proposals of models of computation for AI systems, such as Themodynamic AI \cite{coles2023thermodynamic} or Neuromorphic computing, are required to close the energy gap between the living and non-living.

The field of artificial intelligence (AI) has advanced rapidly in the past few years, and it has given us many tools that are becoming essential for our daily lives. However, training the large models that power these tools is very energy-intensive. Estimates of the CO$_2$ emissions caused by training and using language models are alarming \cite{lacoste2019quantifying, strubell2019energy, faiz2024llmcarbon} and are evidence of the need for more energy-efficient AI systems. Fortunately, multiple deep learning algorithms, such as diffusion models \cite{sohldickstein2015deep, ho2020denoising}, energy models \cite{hopfield1982neural, ackley1985learning, hinton2006fast, lecun2006tutorial}, and many other proposals \cite{liu2023genphys, mate2024neural}, are based on or inspired by physical processes. This relation can be leveraged to create specialized hardware that is better suited for training and running these models. Examples of these approaches include neuromorphic computing \cite{widrow1960adaptive,hopfield1985neural,  mead1990neuromorphic, schuman2017survey, xia2019memristive} which aims to simulate neural networks with hardware that mimics the dynamics of neurons and synapses, physical neural networks \cite{wright2022deep, fu2023photonic, momeni2024training} which use non-conventional physical systems to parametrize and train neural networks, and thermodynamic computing \cite{coles2023thermodynamic, melanson2023thermodynamic}which is a recent proposal that utilizes guided stochastic processes to solve machine learning (ML) problems. There are many other ideas stemming from the classical and quantum domains \cite{mead2012analog, kristensen2021artificial, onen2022nanosecond}. The interplay and inspiration between physical processes and ML will hopefully help improve both the algorithms for AI and the hardware that runs them.

In this paper, we propose a new machine learning algorithm that is based on a simple mechanical system composed of springs and sticks. Previously, the analogy between spring and stick systems and regression models has been discussed rather superficially \cite{levi2023mathematical, loftus2022least}. Here, we build upon this analogy and create an ML algorithm that is universal and has a direct physical system associated with it. The springs and sticks (SS) model is a damped stochastic system that can perform regression over a dataset sampled from a smooth-enough distribution. Dissipative systems have been previously proposed as a way to perform computation \cite{kirkpatrick1983optimization,farhi2001quantum, rebentrost2009environment, albash2018adiabatic}, and we follow this idea to engineer a dissipative universal learning machine. In \cref{sec:math}, we introduce the SS model, construct its Lagrangian, and obtain the equations of motion (EOMs) of such a system. In \cref{sec:code}, we perform simulations of the SS model for regression tasks, and show it can achieve similar performance to that of comparable multi-layer perceptrons (MLPs). In addition, we calculate the system's change in thermodynamic free energy using the Jarzynski equality \cite{jarzynski1997nonequilibrium} and study its relation with the model's performance. We show that the SS model cannot converge to a solution at small physical scales, where environment fluctuations are too large compared to the dynamics of the system. Finally, we conclude with future directions, including possible hardware implementations and interesting applications for the SS model.

\section{The springs and sticks model}
\label{sec:math}

\begin{figure}[!h]
    \centering
    \subfloat[\label{subfig:configuration}]{\includegraphics[width=0.48\textwidth]{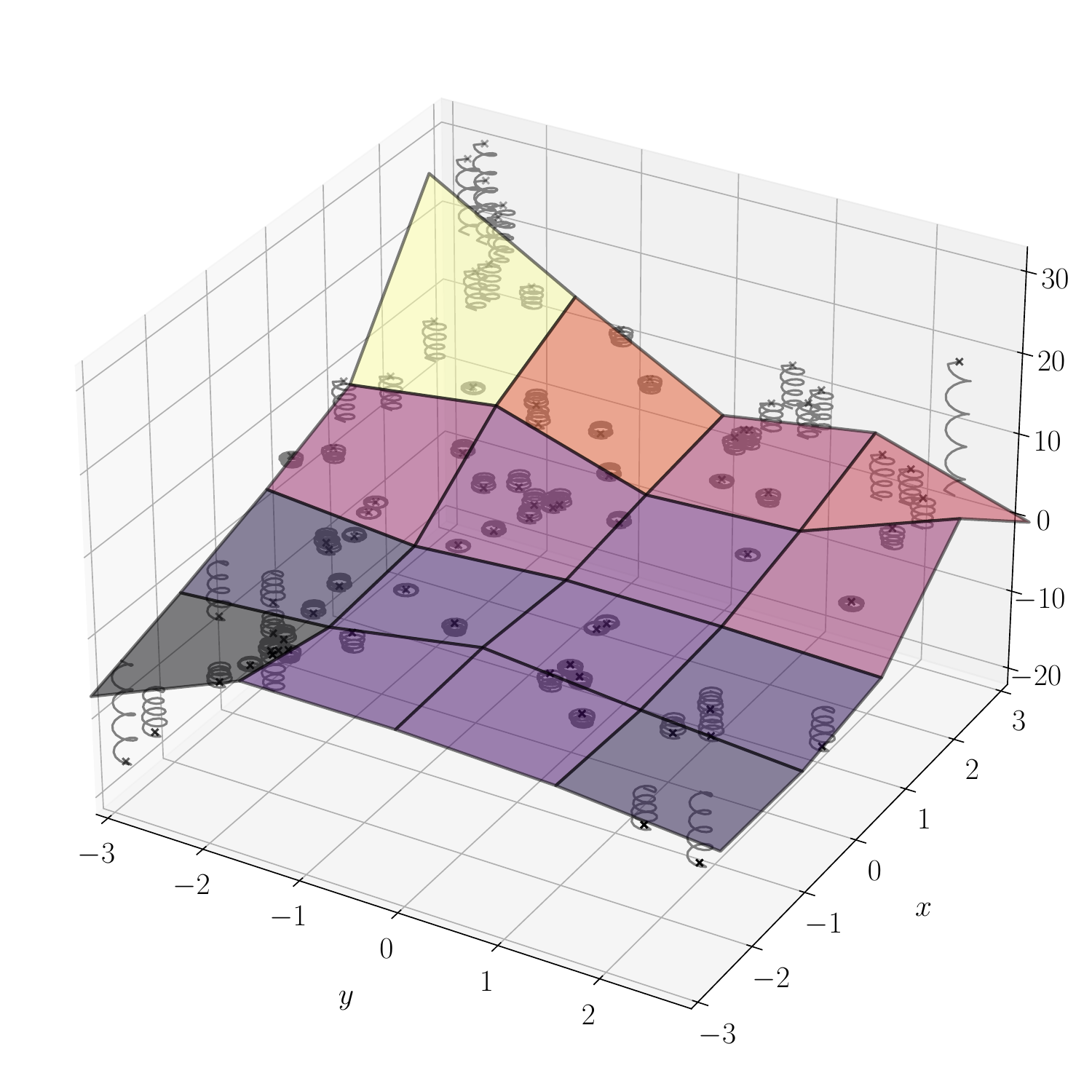}}
    \hfill
    \subfloat[\label{subfig:dynamics}]{\includegraphics[width=0.48\textwidth]{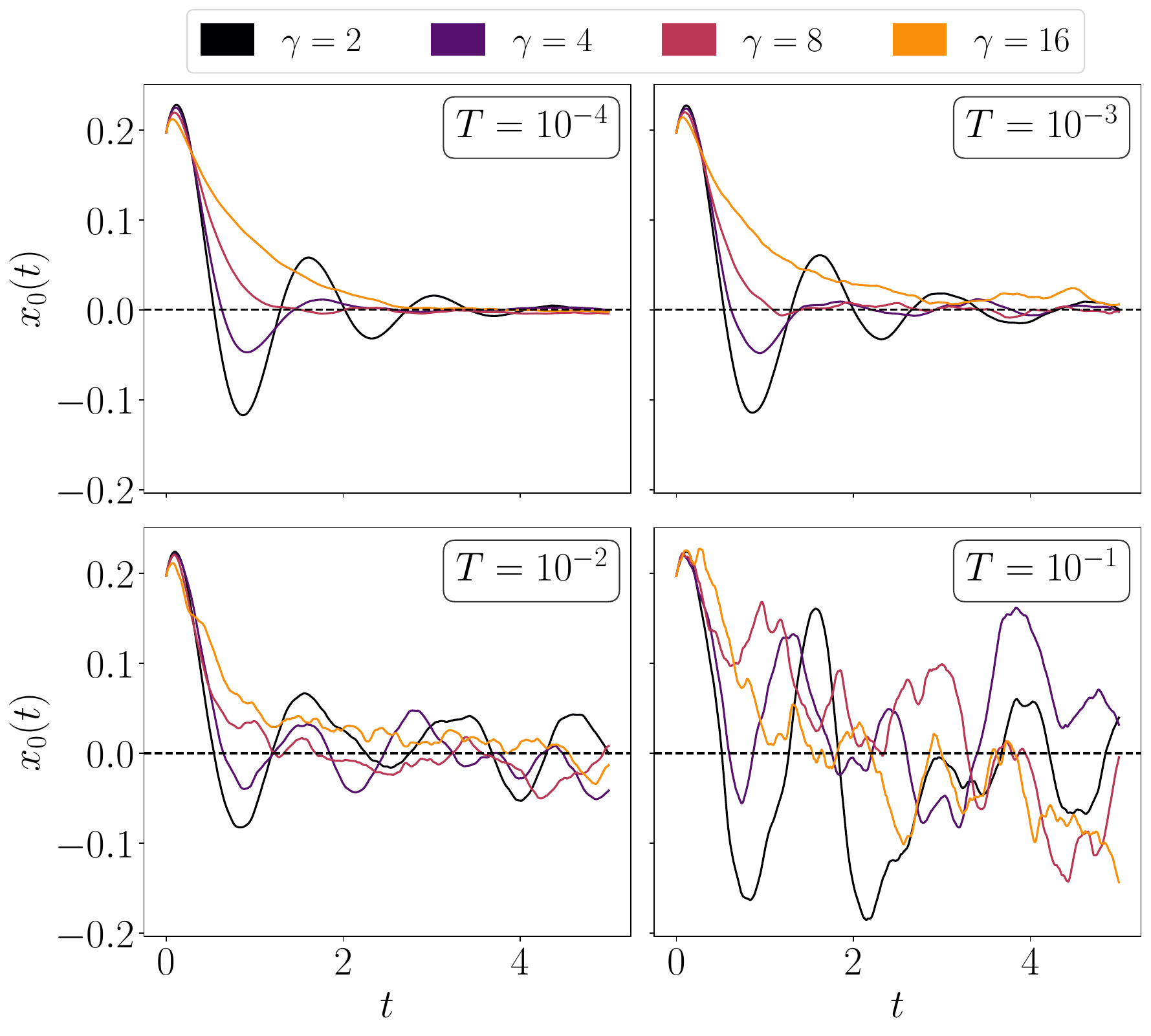}}
    \caption{Simulation of the SS system for simple regression problems. (a) Configuration of the model in a $\mathbb{Z}_{5 \times 5}$ lattice while fitting $f(x,y) = x^2 + x\cdot y^2$ with $80$ datapoints sampled from $(x,y) \sim \mathcal{U}([-\pi, \pi]^2)$ (b) Position of a stick extrema at different time steps for different hyperparameters (friction and temperature) while fitting $f(x)=0$ in a $\mathbb{Z}_{2}$ lattice (one stick). The dashed line indicates the optimal position that minimize the springs' potential energy.}
    \label{fig:system}
\end{figure}

In this section, we define the springs and sticks (SS) model and derive the equations of motion (EOMs) for the system. The SS model, shown in \cref{subfig:configuration}, is a physical system composed of sticks and springs that can be used to solve regression problems. The system is defined on $d+m$ dimensions, where $d$ and $m$ are the dimensions of the input space and output space, respectively. This system is composed of $m\cdot N$ springs and $N_s = \prod_{k=1}^d (N_k - 1)$ sticks, where $N$ is the number of data points and $N_k - 1$ is the number of sticks along dimension $k \leq d$. By attaching springs from the data points to the grid of sticks and dissipating the elastic potential energy, the system relaxes to a minimum-energy configuration that equivalently minimizes a mean-squared error loss. The SS model can be used as a universal approximator of smooth functions via a piecewise-linear fit, and it can be used to solve regression problems on data $\mathcal{D}=\{(\mathbf{u}_i, \mathbf{y}_i ) \}_{i=0}^{N-1}$ where $\dim(\mathbf{u}_i)=d$ and $\dim(\mathbf{y}_i)=m$. This model achieves an approximation error that scales as $O(N_{s}^{-2})$ for any smooth function with finite input and output dimensions. This error is equal to the integration error using the trapezoidal approximation of a function. We discuss universality with more detail and numerically show this error-scaling in \cref{sec:universal}.

\subsection{Lagrangian of a general SS model}

To define the physical system of the SS model, consider a grid of sticks that are connected to the datapoints $\mathcal{D}$ via springs as shown in \cref{subfig:configuration}. The total Lagrangian of this system is the sum of the individual Lagrangians of each stick and spring~\cite{arnol2013mathematical}. A stick in free space has a translational kinetic energy and a rotational kinetic energy of 
\begin{equation}
    K_{\text{tr}}=\frac{1}{2} M \dot{\mathbf{x}}_{\text{com}}^2 = \frac{M}{8}(\dot{\mathbf{x}}_0 + \dot{\mathbf{x}}_1)^2, \quad K_{\text{rot}}=\frac{1}{2} I \omega^2 \approx \frac{M}{24}(\dot{\mathbf{x}}_1 - \dot{\mathbf{x}}_0)^2
\end{equation}
where ${\mathbf{x}}_{\text{com}}$ is the center of mass, $\mathbf{x}_0$ and $\mathbf{x}_1$ are the extrema of the stick, and $I$ is its moment of inertia. Thus, for a grid of sticks where the extrema of neighboring sticks are held together, the total kinetic energy is $K = \sum_{\text{sticks}} K_{\text{tr}} + K_{\text{rot}}$ since the kinetic energy is invariant under euclidean translations. In addition, the potential energy of a spring with rest length $\ell_0 = 0$ and spring constant $k$ is given by $U = \frac{1}{2} k \delta^2$, where $\delta$ is the displacement of the spring. Thus, attaching $m$ springs between two points $(\mathbf{u}_i, y^j_i)$ and $(\mathbf{u}_i, \hat{y}^j)$, one spring per coordinate $j\in[m]$, yields a potential energy of $U_i = \frac{1}{2} k \sum_{j=0}^{m-1} ( \hat{y}^j(\mathbf{u}_i) - y^j_i )^2$ per data point. Here, $\hat{y}^j (\mathbf{u}_i)$ is the $j$-th coordinate of the predicted target of an input $\mathbf{u}_i$, and we will later define it for the SS model. Using these energies, we define the Lagrangian of a general SS system. For a grid $\mathbb{Z}_{\vec{N}}=\mathbb{Z}_{N_1} \times \cdots \times \mathbb{Z}_{N_d}$ of $\prod_{k=1}^d (N_k-1)$ sticks and $m \cdot N$ springs we have
\begin{equation}
\begin{aligned}
    \label{eq:sslagrangian}
    L &= K - U  \\
    &= \frac{M}{8} \sum_{\vec{i} \in \mathbb{Z}_{\vec{N}}} \sum_{b=0}^{d-1} \sum_{p=0}^{m-1}  \left( \dot{x}^p_{\vec{i} + \hat{e}^b} + \dot{x}^p_{\vec{i}} \right)^2 + \frac{M}{24} \sum_{\vec{i} \in \mathbb{Z}_{\vec{N}}} \sum_{b=0}^{d-1} \sum_{p=0}^{m-1}  \left( \dot{x}^p_{\vec{i} + \hat{e}^b} - \dot{x}^p_{\vec{i}} \right)^2  
    \\& \hspace{8cm}  - \frac{k}{2} \sum_{j=0}^{N-1} \sum_{p=0}^{m -1} \left( \hat{y}^{p}(\mathbf{u}_j) - y^{p}_{j} \right)^2,
\end{aligned}
\end{equation}
where $\hat{e}^b$ is the unit vector in the $b$-th direction and $x_{\vec{i}}$ is the position of the sticks-grid at point $\vec{i}$. Depending on the boundary terms of \cref{eq:sslagrangian}, the grid of sticks can have dangling edges or not---meaning that the two extrema of each stick is attached to other sticks instead of being detached. For simplicity, we consider grids with no dangling edges.

It is worth noting that this system suffers from \emph{the curse of dimensionality}, as the amount of dynamical variables $\nu := |\mathbb{Z}_{\vec{N}}|$ grows exponentially in the dimension of the domain $d$. This issue can be mitigated with divide-and-conquer approaches \cite{jacobs1991adaptive, bungartz2004sparse, cormen2022introduction}. For example, multiple SS models can be "trained" on fewer features and the predicted target is then post-processed. In such case, the amount of dynamical variables would instead scale as $\nu = O(\kappa C)$ where $\kappa$ is the number of SS models and $C \leq N_k^{d'}$ is a fixed constant that depends on the largest domain dimension $d'$ from the SS sub-systems. Similarly, to reduce the number of springs used when training on a large dataset, we can use a mini-batch approach as done with neural network training. In such case, we time-evolve the SS system for a short time $\Delta t$ with a potential energy corresponding to a mini-batch of constant size $B < N$, reducing the amount of springs needed to $O(mB)$.

For simplicity, to perform inference with the SS model, we have assumed that the positions of extrema of every stick $(\mathbf{r}(\vec{i}), \mathbf{x}_{\vec{i}})$ only vary in the last $m$ dimensions and is fixed in the first $d$ dimensions. We then choose $\hat{\mathbf{y}}$ to be a linear combination of the extrema positions of the sticks enclosing an input $\mathbf{u}$:
\allowdisplaybreaks
\begin{equation}
\label{eq:inference}
\begin{aligned}
    \hat{y}^p (\mathbf{u}) &= \sum_{\vec{i} \in \mathbb{Z}_{\vec{N}}} \mathbf{1}_{\left \{\mathbf{u} \in \Omega_{\vec{i}} \right \}}\sum_{\vec{j} \in \{0,1\}^d} x^p_{\vec{i}+\vec{j}} \prod_{b=0}^{d-1} \left[ (1-\uplambda^b_{\vec{i}}(\mathbf{u})) \delta_{j^b,0} + \uplambda^b_{\vec{i}}(\mathbf{u}) \delta_{j^b,1} \right] \\
    \Omega_{\vec{i}}&:=\prod_{b=0}^{d-1} [\mathbf{r} ({\vec{i}}), \mathbf{r}({\vec{i}+\hat{e}^b}))\\
    \uplambda_{\vec{i}}(\mathbf{u}) &:=(\mathbf{u} - \mathbf{r}(\vec{i}))\odot \vec{\ell}^{-1} = (\mathbf{u} - \vec{i} \odot \vec{\ell} - \mathbf{r}(\vec{0}))\odot \vec{\ell}^{-1},
\end{aligned}
\end{equation}
where $\Omega_{\vec{i}}$ is the cell in the mesh with origin at $\vec{i}$, $\uplambda_{\vec{i}}(\mathbf{u})$ quantifies how far $\mathbf{u}$ is from the grid point $\vec{i}$, $\vec{\ell}$ is the sticks-grid spacing in each direction, and $\vec{\ell}^{-1}$ is the element-wise inverse of $\vec{\ell}$. These assumptions of having fixed $\mathbf{r}(\vec{i})$ and lengths $\vec{\ell}$ can be dropped to obtain a more expressive SS model. For example, with variable lengths, the piecewise-linear approximation achieved will behave more like that of a standard multi-layer perceptron (MLP) with one hidden layer and ReLU activations (combination of piecewise-linear functions with variable length). For more intuition on the SS model, we discuss simplified examples for 1D functions in \cref{sec:simple_cases}. 

\subsection{Equations of motion and convergence}

The non-dissipative EOMs of the SS system can be obtained by solving the Euler-Lagrange equations, which we discuss in \cref{sec:eulerlagrange}. However, for the system to converge to a solution of the regression problem we need energy dissipation, otherwise the springs and sticks will keep oscillating indefinitely. To introduce dissipation we use a Langevin thermostat. We add friction terms to the EOMs which are equivalent to submerging the system in a thermal bath. The system's EOMs then become a linear Langevin equation of the form
\begin{equation}
    \frac{d}{dt} \begin{pmatrix}
    \mathbf{x} \\
    \mathbf{\dot{x}}
    \end{pmatrix} = \mathbf{A} \begin{pmatrix}
        \mathbf{x} \\
        \mathbf{\dot{x}}
        \end{pmatrix} + \mathbf{b} + \mathbf{B} \dot{\mathbf{\xi}} (t),
\end{equation}
where $\mathbf{A}$ and $\mathbf{b}$ are dictated by the mass matrix $\mathbf{M}$, the force vector $\mathbf{f}(\mathbf{x}, \mathbf{\dot{x}},t)$, and the friction factor $\gamma$ of the system, and $\mathbf{\xi}(t)$ are $w$ independent standard Wiener processes that interact with $(\mathbf{x}, \mathbf{\dot{x}})$ via $\mathbf{B}\in M_{2\nu \times w} (\mathbb{R})$. The stochastic noise term $\mathbf{B} \dot{\mathbf{\xi}} (t)$ is controlled by the temperature of the system and is required to respect the fluctuation-dissipation theorem~\cite{callen1951irreversibility, kubo1957statistical, seifert2012stochastic, mishin2016energy, peliti2021stochastic}. These fluctuations help the system escape local minima, similar to the role of stochastic gradient descent~\cite{robbins1951stochastic, bottou2018optimization, kovachki2021continuous}. 

For simplicity, we set $w=2\nu$ and $\mathbf{B}$ to be diagonal. This allows us to write the EOMs of the SS model as
\begin{equation}    
    \begin{aligned}
        \frac{d}{dt} \mathbf{x} &= \mathbf{\dot{x}}, \\
        \frac{d}{dt} \mathbf{\dot{x}} &= {\mathbf{M}}^{-1} \mathbf{f}(\mathbf{x}, \mathbf{\dot{x}},t) - \gamma \mathbf{\dot{x}} + \sigma \dot{\mathbf{\xi}} (t).
    \end{aligned}
\end{equation}
where $\sigma = \sqrt{2 \gamma T k_b / M}$ is the standard deviation of the noise term, $T$ is the temperature of the system, $k_b$ is the Boltzmann constant, $M$ is the mass of a stick, and $\gamma$ is in units of inverse time. These dissipative dynamics can be solved numerically using the Euler-Maruyama method \cite{protter2005stochastic, karatzas2014brownian, li2020scalable}. For different values of $\gamma$, $k$, and $M$, the dynamics can be underdamped, overdamped, or critically damped. A simple example of these dynamics are shown in \cref{subfig:dynamics}. Overall, this system can be understood as a machine learning model that diffuses the parameters of the model to the optimal parameter configuration, similar to the recent proposal in Ref. \cite{wang2024neuralnetworkdiffusion}.

Given the linearity of this system, we can study relevant physical quantities such as its relaxation time ($\tau_r \propto \norm{\mathbf{A}}^{-1}$) and its entropy production \cite{spinney2012entropy, landi2013entropy} (for details see \cref{sec:entropyProd}). Moreover, convergence speed bounds for this model can be found using stochastic thermodynamics \cite{seroussi2023speed}. In this system, when $k=0$ and $M>0$, $\mathbf{f}$ vanishes and the system's configuration $\mathbf{x}$ evolves as a free Brownian particle, so $\operatorname{Var}[\mathbf{x}] \propto t$ grows linearly with time. On the other hand, when $k>0$ and $M=0$, the system's configuration $\mathbf{x}$ evolves as a Ornstein–Uhlenbeck process, and thus $\operatorname{Var}[\mathbf{x}]\propto k_b T/k$ saturates to a constant when $t\rightarrow \infty$. This suggests that the SS system will have problems learning in some cases, such as when $k\ll k_b T$. We study this phenomenon in \cref{sec:code}.

\section{Numerical experiments}
\label{sec:code}

In this section, we simulate the EOMs of the system described in \cref{sec:math}. First, we consider a 2D regression problem where the data is generated from an analytical function. We compare the performance of the SS model with that of simple multi-layer perceptrons (MLPs). Then, we find a relation between the free energy released by the system and its ability to learn an underlying distribution, and empirically show that a minimum free energy $\Delta F_{\text{min}}$ is required for the SS system to learn. We believe this is a general property of physical systems that learn from data and provide arguments in favor of this claim.

\subsection{Training SS models on simple datasets}

\begin{figure}[!h]
    \centering
    \includegraphics[width=1\textwidth]{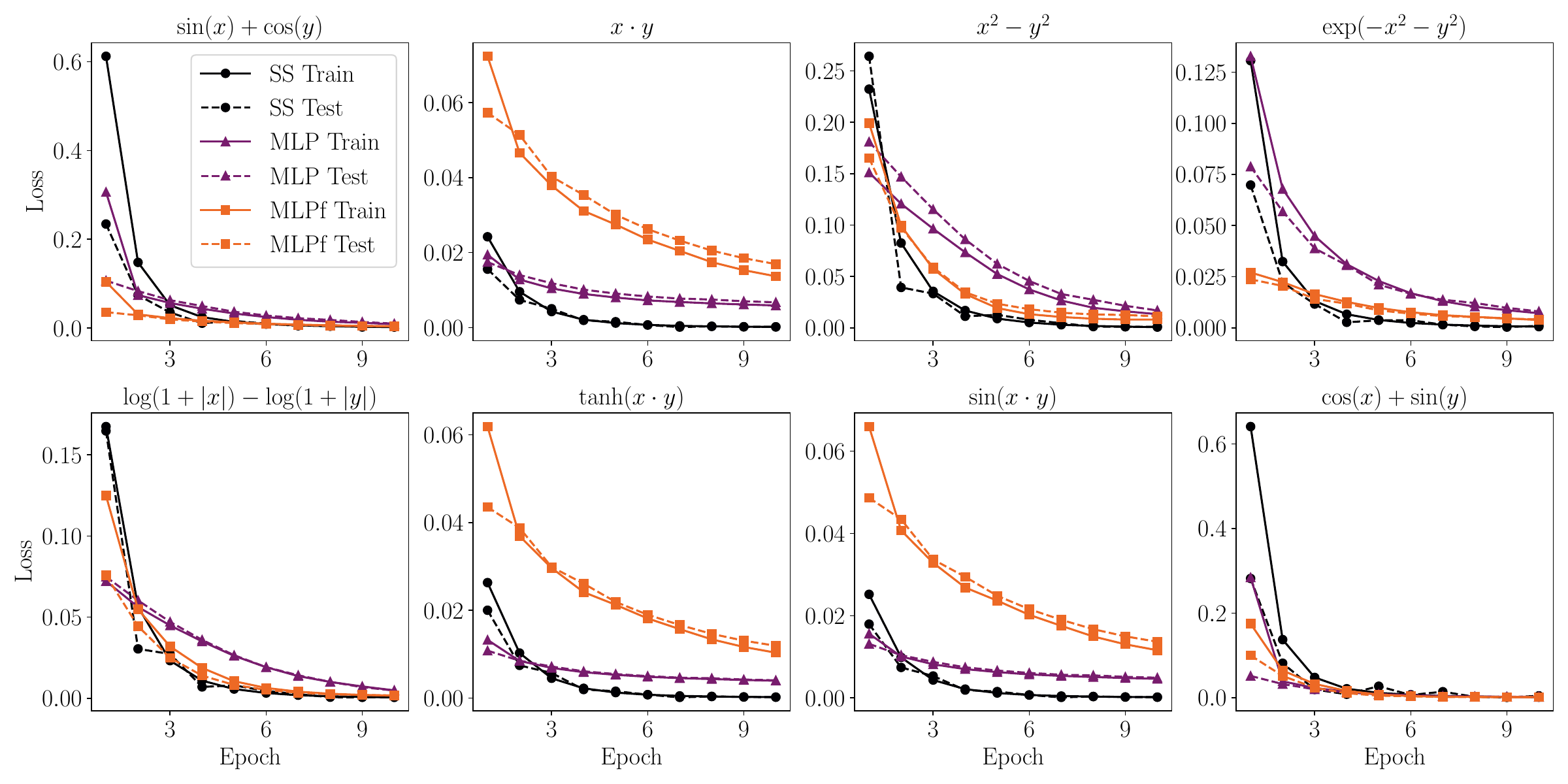}
    \caption{Comparison of the loss of the SS model and single-layer MLPs for regression on data sampled from smooth $f:\mathbb{R}^2 \longrightarrow \mathbb{R}$ functions perturbed with a small Gaussian noise $\epsilon \sim 0.01 \cdot \mathcal{N}(0,1)$. All models are trained on the same $160$ datapoints uniformly sampled from $[0,1]^2$ and equal batch size of $16$ every epoch. The chosen hyperparameters are $\gamma=10$, $M=1$, $k=1$, and $T=10^{-3}$. The SS model has $4 \times 4$ sticks and the MLPs have one hidden layer with $16$ neurons and ReLU activations. In the legend, the weights and biases of the MLP are trainable, while the biases of the MLPf are fixed to a random value (similar to having sticks with non-variable length).}
    \label{fig:comparison}
\end{figure}

We simulate and train the SS model on synthetic data generated from analytical functions $f:\mathbb{R}^2 \longrightarrow \mathbb{R}$. We set the length of each stick and the number of sticks per dimension to be constant. We use a $4\times 4$ grid composed of $N_s=16$ sticks and train on $N=160$ points. The initial positions of the sticks-grid are randomly initialized. In each epoch we time-evolve the SS system for $\Delta t = 0.1$, use two time steps ($0$ and $\Delta t$) in the Euler-Maruyama integral, and use a batch of $16$ datapoints. These choices allow the system to train in faster time and compares well to the optimization of a loss function using gradient descent. We compare the loss of this SS model with that of single-layer MLPs and find that they achieve similar performance as shown in \cref{fig:comparison}. The speed of convergence is mainly affected by the friction coefficient, and when it is high enough, the system can converge in very few epochs. In these examples, the SS model can achieve better performance than the MLPs, but it must be noted that it also takes a longer time to run per epoch. One of the reasons being that our code has not been optimized for speed.

\subsection{Thermodynamic learning barriers}

Initializing the SS system requires some input energy that, in the absence of thermal fluctuations, depends on the number of springs (size of the dataset) and spring constant $k$. The system then dissipates energy due to interactions with the bath and equilibrates to a Boltzmann distribution. If the system is too small ($k \sim k_b T$) the thermal fluctuations become larger with respect to the system's motion as shown in \cref{subfig:dynamics}. These large fluctuations lead to a situation where, on average, the system does not dissipate a meaningful amount of free energy and cannot learn the underlying distribution accurately. The energy dissipated by the system and the ability to learn the underlying distribution are related, since no energy dissipation implies no convergence to a solution. Moreover, when learning is achieved, the entropy of the final distribution must be smaller than that of the initial distribution. Thus, heat must be released to the environment similar to Landauer's principle \cite{landauer1961irreversibility, bennett1982thermodynamics, bennett2003notes}, which states that the energy of erasing one bit of information is $E \geq k_b T \ln 2$. This entropy decrease is independent of the underlying ML model if one associates one set of parameters of the model to a point in the phase space of the physical system realizing it.

\begin{figure}[!h]
    \centering
    \includegraphics[width=0.7\textwidth]{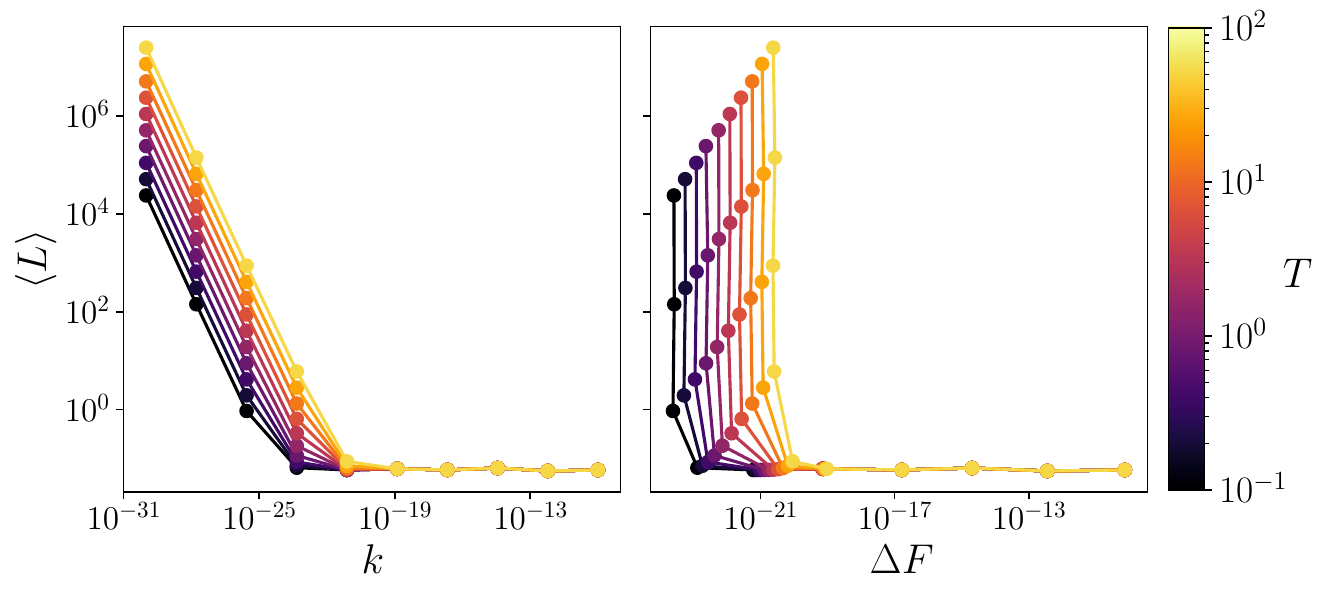}
    \caption{Average loss in a steady state solution as a function of "scale" and of free energy. The free energy is calculated for each value of $k$ shown. We set the mass $\abs{M}=\abs{k}$, the friction coefficient $\gamma = 0.1$, the number of datapoints $N=20$, the number of sticks $N_s = 1$, and the underlying distribution $f(x)=\cos(x)$ with $x \sim \mathcal{U}([0, 2\pi])$.  The behaviour is similar for other values of $N_s$, $N$, and $f$.}
    \label{fig:loss_scale}
\end{figure}

Understanding the relation between energy and the learning process of a model can guide the design of better hardware and software for ML. Previous work has focused on understanding the learning efficiency of neural networks~\cite{goldt2017stochastic}. Here, to understand the energy-learning relation of SS models, we compute the free energy change and the average training loss of the system once it reaches a steady state solution. We consider the average loss instead of the loss of the average parameters obtained from multiple parameter samples in the steady state solution, since this averaging process requires energy dissipation too. We calculate the change in free energy\footnote{We use the convention of $\Delta F = F_{i} - F_{f}$ to measure the energy dissipated to the environment, rather than $\Delta F = F_{f} - F_{i}$ which quantifies the energy change of the system.} via the Jarzynski equality
\begin{equation} 
\Delta F = F_{i} - F_{f} = k_b T \ln \langle e^{-W/k_b T} \rangle,
\end{equation}
where $W$ is the work done on the system and the average is taken over multiple evolution trajectories \cite{jarzynski1997nonequilibrium}. This free-energy difference is taken between the initial configuration (randomly initialization) and the final configuration of the sticks-grid (after training). We plot these two quantities for multiple "scales" (i.e. different values of $k$ and $M$) and different temperatures in \cref{fig:loss_scale}. In addition, given the linearity of the system, we can calculate other relevant thermodynamic quantities, such as the entropy production rate and flux rate of the system with analytical expressions \cite{landi2013entropy} as shown in \cref{sec:entropyProd}.

\begin{figure}[!h]
    \centering
    \subfloat[\label{subfig:df_expressivity}]{\includegraphics[width=0.45\textwidth]{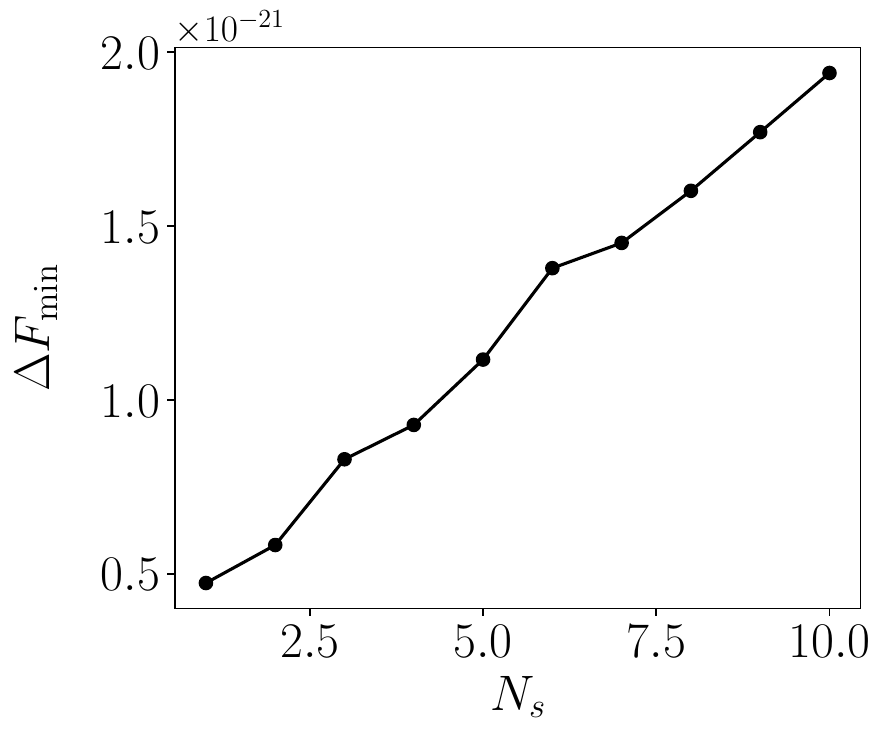}}
    \hfill
    \subfloat[\label{subfig:df_heatmap}]{\includegraphics[width=0.5\textwidth]{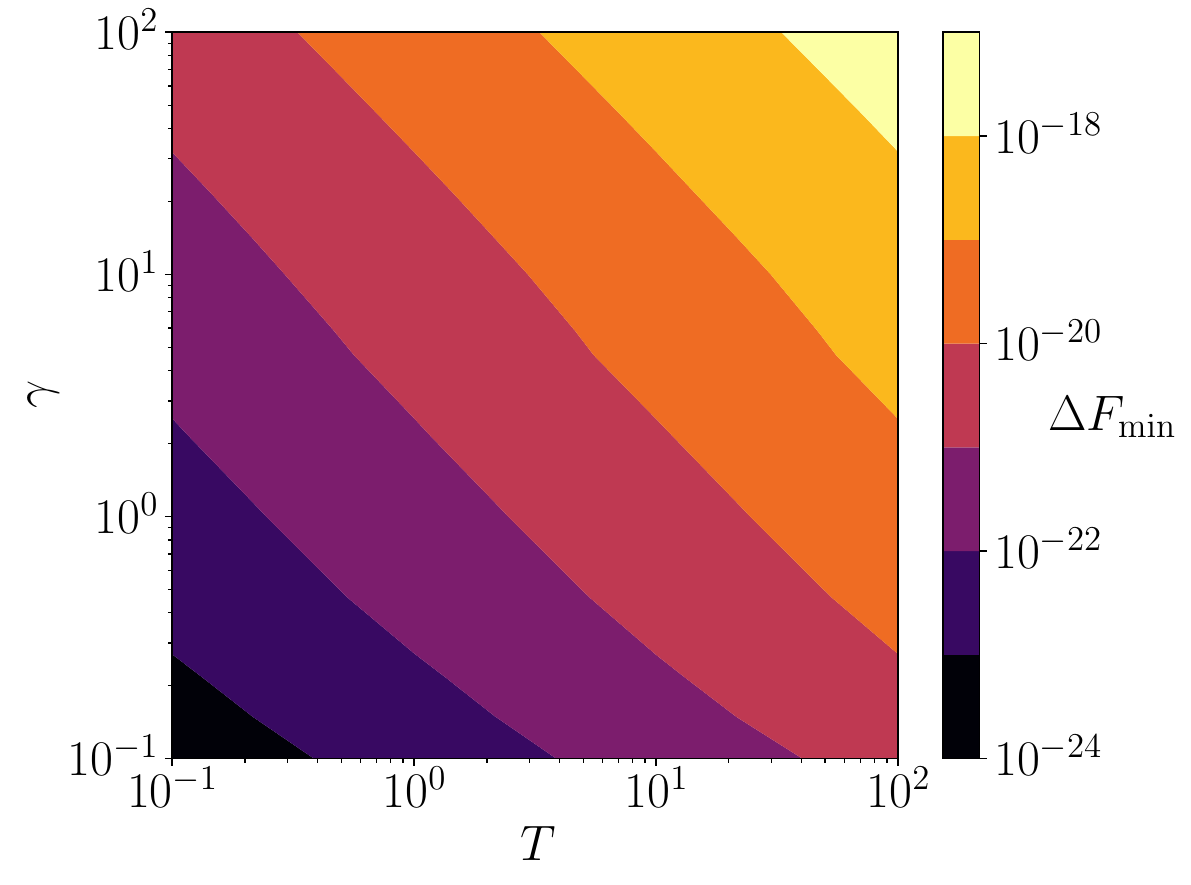}}
    \caption{TLB as a function of different model parameters in SI units. (a) TLB as a function of the expressivity of the model (number of sticks $N_s$) with $\gamma=1$ and $T=1$. (b) Contour plot of the TLB as a function of the friction coefficient $\gamma$ and the temperature $T$. All other parameters are the same as those from \cref{fig:loss_scale}.}
    \label{fig:system}
\end{figure}

We observe that as we make the system smaller, this is, as we decrease $k$ and $M$, the free-energy released by the system decreases until it plateaus at a minimum value $\Delta F_{\text{min}}$ (see \cref{subfig:df_heatmap}). In such cases, the SS system is uncapable of dissipating energy and therefore of learning. This is reflected in the rapid increase of the average loss. We call this minimum free-energy value a \emph{thermodynamic learning barrier} (TLB). Similar to Landauer's principle, we consider this bound $\Delta F_\text{min} \leq \Delta F$ to be the minimum energy required to learn from a given dataset with an SS model. We calculate the TLB for different expressivities (i.e. number of sticks) in \cref{subfig:df_expressivity} and different values of the friction coefficient and temperature in \cref{subfig:df_heatmap}.

We find that in this system the TLB is proportional to the expressivity of the SS model. This intuitively aligns with Landauer's principle, where $\Delta S \propto \ln V$ and $V$ is the volume in phase space. Then, since $V \propto (x \cdot p)^{N_s}$ we know $\Delta S \propto N_s$. We conjecture that this relation relates to the scaling laws in deep learning \cite{hestness2017deep, kaplan2020scaling, bahri2024explaining}, where the optimal loss---determined by the expressivity of the model---follows a negative power-law with respect to the compute power and the number of parameters in the model. In particular, given the relation between $N_s$ and the approximation error $E$ (see \cref{sec:universal}), the SS model has a similar power-law relation between the approximation error and the compute power, $E = O(\Delta F_{\text{min}}^{-2})$. We also find that the TLB is proportional to the temperature and the friction coefficient, $\Delta F_{\text{min}} \propto k_b T \gamma$. That is, at higher temperature and higher friction the system requires a larger energy input to learn an underlying data distribution. The behaviour of the TLB is similar for different underlying functions $f$. These barriers quantify the minimum energetic requirements of the SS model for learning functions from data. We believe this behavior is common in other machine learning models and hope TLBs help benchmark and develop better energy-efficient hardware.

% two subfigures side by side

\section{Conclusion}

In this work, we have proposed a physical system composed of springs and sticks capable of learning an arbitrary approximation of any smooth function. We have shown that this system can be used to solve regression problems on simple datasets and its performance is comparable to that of multi-layer perceptrons (MLPs) with one single hidden layer. We have also shown that the system has a minimum energetic requirement to learn from a given dataset. We call this minimum energy a thermodynamic learning barrier (TLB), and show it is proportional to the expressivity of the model, the temperature of the environment, and the friction coefficient of the system. We believe this simple linear Langevin system helps us better connect ideas from machine learning and physics. 

% The code used to generate the results in this work has not been significantly optimized for speed.

Interesting research directions include extending the SS model to "deep" SS models similar to deep neural networks, using a different loss function via a different potential energy such as the cross-entropy loss, and doing an in-depth analysis of the TLBs and their relation to neural scaling laws. In addition, we believe these dynamics can be mapped to an electronic circuit as studied in Ref.~\cite{panuluh2017lagrangian}, allowing for the development of specialized hardware to train SS models or to implement such models in a thermodynamic processing unit~\cite{coles2023thermodynamic}. Another interesting idea to explore is the implementation of the SS model in a quantum computer. For example, using the algorithm proposed in Ref.~\cite{babbush2023exponential} which shows an exponential speed-up for solving the dynamics of a coupled springs system, or using quantum algorithms that solve differential equations~\cite{berry2014high, Childs_2017, berry2017quantum, costa2023further, bagherimehrab2023fast} to simulate the dynamics of the SS system.

% Future work can be doing dynamics with Hessian instead of following the system's dynamics (second order optimizations)

% Cooling of quantum system \cite{xu2014demon}

% One can map a ReLU network with one hidden layer to a SS model with variable lengths. This can inspire analyzing the dynamics of ReLU networks with a physical system.

\subsection*{Acknowledgments}

The authors acknowledge helpful feedback from Lasse Bj{\o}rn Kristensen, Philipp Schleich and Austin Cheng, and early discussions with Matteo Marsili and Dvira Segal. This work is supported by the Novo Nordisk Foundation, Grant number NNF22SA0081175, NNF Quantum Computing Programme. A.A.-G. thanks Anders G. Fr{\o}seth for his generous support. A.A.-G. also acknowledges the generous support of Natural Resources Canada and the Canada 150 Research Chairs program.

\vspace{0.5cm}
\noindent \textbf{Code availability:} Code, figures, and animations are available at \href{https://github.com/bestquark/springs-and-sticks}{https://github.com/bestquark/springs-and-sticks}.

% \nocite{*}
% \bibliographystyle{unsrt}
% \bibliography{bibliography}
\printbibliography

\newpage
\appendix

\centerline{\Large \textbf{Supplementary Material}}

\section{Universal approximation}
\label{sec:universal}

\begin{figure}[!h]
    \centering
    \includegraphics[width=0.7\textwidth]{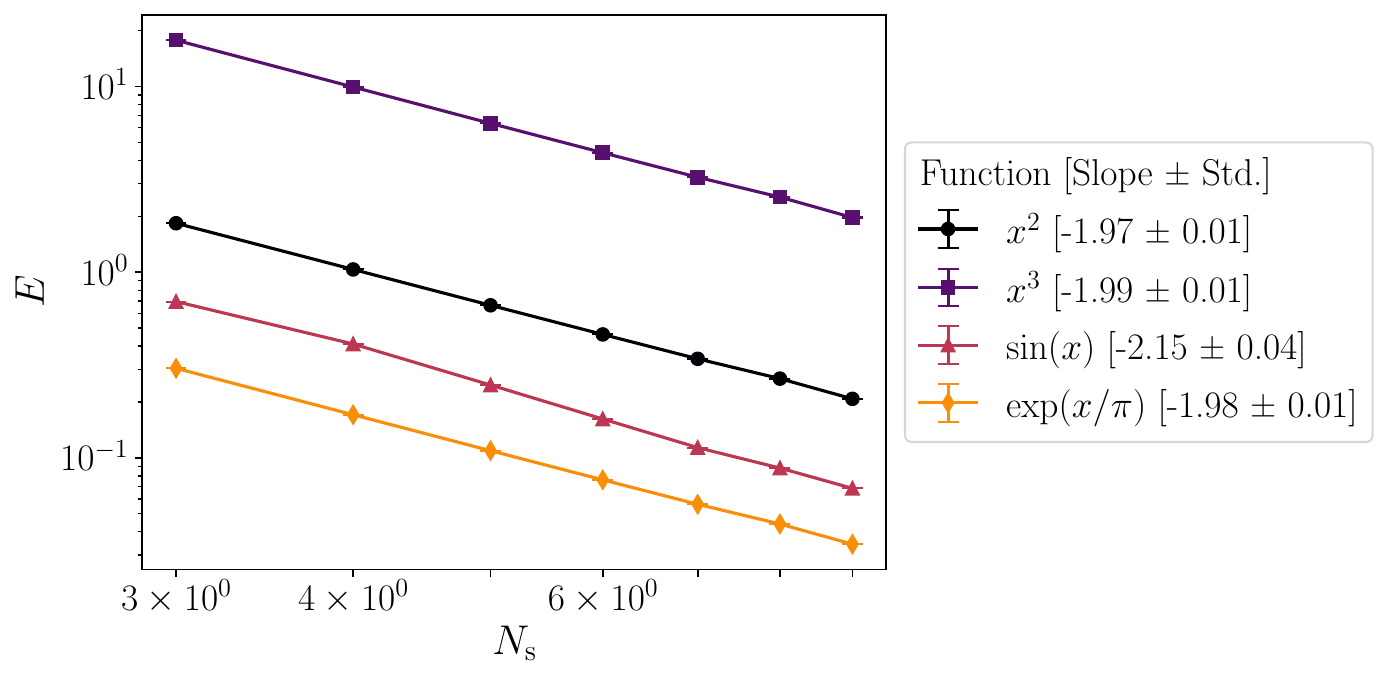}
    \caption{Approximation error of the springs and sticks system for four smooth functions in the domain $x\in[0,2\pi)$ as a function of the number of sticks. Each point is the average of 8 runs with different random initial conditions. We attribute the lower-than $2$ slope in some functions to the system being in a non-zero temperature state.}
    \label{fig:error_analysis}
\end{figure}

The springs and sticks system is a universal approximator of continuous functions with bounded Hessian. This is simple to see given that the system can implement a piecewise-linear approximation of a function with arbitrary precision. For simplicity, consider a mesh with $\vec{\ell}=(\ell, \dots, \ell)$ and $N_1 = \cdots = N_d = N$, comprised of multiple $d$-dimensional polygons $\Omega_i$. The error of such approximation with respect to a function $f:[0,1]^d \subseteq \mathbb{R}^d \longrightarrow \mathbb{R}$ can be upper bounded similar to the Trapezoidal rule. When $x_{\vec{i}} = f(\mathbf{r}(\vec{i}))$ for all $\vec{i} \in \mathbb{Z}_{\vec{N}}$, the error of the approximation is given by 
\begin{equation}
\begin{aligned}
    E&= \sum_{\vec{i} \in \mathbb{Z}_{\vec{N}}} \int_{\Omega_i} \abs{f(\mathbf{u}) - \hat{y}(\mathbf{u})}^2 d\mathbf{u}\\
    &= \sum_{\vec{i} \in \mathbb{Z}_{\vec{N}}} \int_{\Omega_i} \abs{f(\mathbf{u}) - \sum_{\vec{j} \in \{0,1\}^d} f(\mathbf{r}(\vec{i}+\vec{j})) \prod_{b=0}^{d-1} \left[ (1-\uplambda^b_{\vec{i}}(\mathbf{u})) \delta_{j^b,0} + \uplambda^b_{\vec{i}}(\mathbf{u}) \delta_{j^b,1} \right]}^2 d\mathbf{u}.
\end{aligned}
\end{equation}
For a function $f$ with bounded Hessian, this approximation error can be upper bounded by $E \leq \mathcal{O}\left( N^{-2} \right)$ \cite{bungartz2004sparse}. This approximation error is independent of the dimension $d$. A simple error analysis is shown in \cref{fig:error_analysis} for different functions $f:\mathbb{R} \longrightarrow \mathbb{R}$, where the error decreases as the number of sticks increases with a slope $m\approx -2$.

\section{Simple SS models}
\label{sec:simple_cases}

\subsection*{Linear 1D function $f:\mathbb{R} \longrightarrow \mathbb{R}$}

A system of a stick (rod) coupled to several springs can resemble a linear fitting of a dataset $\mathcal{D}=\{(u_i, y_i)\}_{i=0}^{N-1}$. This system was considered in \cite{levi2023mathematical} as means for solving a 1D linear regression. We will set $\ell = \max \{ u_i \}_{i=0}^{N-1} - \min \{ u_i \}_{i=0}^{N-1}$ the length of the stick and $x_0$ and $x_1$ the vertical position of the extrema of the rod. In the limit of small oscillations the approximation $\arcsin((x_1 - x_0)\ell^{-1}) \approx (x_1 - x_0)\ell^{-1} $ holds. For this regime, we can write the Lagrangian of the system as
\begin{equation}
L = \underbrace{\frac{M}{8} \left(\dot{x}_0 + \dot{x}_1 \right)^2\vphantom{\Bigg|}}_{K_{\text{tr}}} + \underbrace{\frac{M}{24} \left( \dot{x}_1 - \dot{x}_0 \right)^2\vphantom{\Bigg|}}_{K_{\text{rot}}} -  \underbrace{\frac{k}{2}\sum_{i=0}^{N-1} \left(x_0 \left(1 - \uplambda_i \right) + x_1 \uplambda_i - y_i \right)^2\vphantom{\Bigg|}}_{U},
\end{equation}
where $\uplambda_i = (u_i-u_0)\ell^{-1}$. For this system, we use $1$ stick and $N$ springs.

\subsection*{Non-linear 1D function $f:\mathbb{R} \longrightarrow \mathbb{R}$}

A generalization of the previous case is a system that uses $N_s$ sticks connected via pivot joints. For simplicity, we will take them to be of equal length such that $\ell N_s = \max \{ u_i \}_{i=0}^{N-1} - \min \{ u_i \}_{i=0}^{N-1}$. Thus, we have $N_s + 1$ degrees of freedom $\{ x_0, x_1, \dots, x_{N_s} \}$ where $(x_j, x_{j+1})$ are the vertical position of the extrema of stick $j$. Using the small angle approximation, we can write the Lagrangian of the system as $L = K_{\text{tr}} + K_{\text{rot}} - U$, where
\begin{equation}
    \begin{aligned}
        K_{\text{tr}} &=\frac{M}{8} \sum_{i=0}^{N_s -1} \left( \dot{x}_{i+1} + \dot{x}_i \right)^2, \\
        K_{\text{rot}} &= \frac{M}{24} \sum_{i=0}^{N_s -1} \left( \dot{x}_{i+1} - \dot{x}_i \right)^2,  \\
        U &= \frac{k}{2}  \underset{u_j \in [i\ell, (i+1)\ell]}{\sum^{N_s -1}_{i=0} \sum^{N -1}_{j=0 }} \left( x_i + \left( u_j - i\ell - u_0 \right) \frac{x_{i+1} - x_{i}}{\ell} - y_j \right)^2 \\
        &= \frac{k}{2} \sum_{i=0}^{N_s -1} \sum_{j=0}^{N -1} H(\uplambda_{i}(u_j) ) H(-\uplambda_{i+1}(u_j) ) \left( x_i + \uplambda_{i}(u_j)  (x_{i+1} - x_{i}) - y_j \right)^2,
    \end{aligned}
\end{equation}
where $\uplambda_{i}(u_j) = \left( u_j - i\ell - u_0 \right)\ell^{-1}$ and $H$ is a Heaviside step function used to define the cell $\Omega_{\vec{i}}$ from \cref{eq:inference}. Here $\ell$ is chosen such that the piecewise-linear approximation of $f$ with width $\ell$ is close to $f$ up to a desired error $\varepsilon$. This system requires $N_s={O}(\varepsilon^{-1/2})$ sticks and $N$ springs.

\section{Euler-Lagrange equations}
\label{sec:eulerlagrange}
To obtain the EOM of the system, we need to solve the Euler-Lagrange equations for the Lagrangian of \cref{eq:sslagrangian}. These are given by
\begin{equation}
     \frac{\partial L}{\partial x^p_{\vec{i}}} - \frac{d}{dt} \frac{\partial L}{\partial \dot{x}^p_{\vec{i}}} = 0.    
\end{equation}
The first term only depends on the potential energy $U$, 
\begin{equation}
\begin{aligned}
    \label{eq:partial_x}
    \frac{\partial L}{\partial x^p_{\vec{i}}} &= -k \sum_{j=0}^{N-1} \left( \hat{y}^p(\mathbf{u}_j) - y^p_{j} \right) \frac{\partial \hat{y}^p(\mathbf{u}_j)}{\partial x^p_{\vec{i}}} \\
    & = -k \sum_{j=0}^{N-1} \left( \hat{y}^p(\mathbf{u}_j) - y^p_{j} \right) \sum_{\substack{
        \vec{v} \in \mathbb{Z}_{\vec{N}} \\ \vec{t} \in \{0,1\}^d}} \mathbf{1}_{\left \{\mathbf{u}_j \in \Omega_{\vec{v}} \right \}} \delta_{\vec{i}, \vec{v}+\vec{t}} \  \cdots  \\
        & \hspace{3cm} \cdots \prod_{b=0}^{d-1} \left[ (1-\uplambda^b_{\vec{v}}(\mathbf{u}_j)) \delta_{t^b,0} + \uplambda^b_{\vec{v}}(\mathbf{u}_j) \delta_{t^b,1} \right],  
\end{aligned}
\end{equation}
and the second term only depends on the kinetic energy $K$, 
\begin{equation}
    \begin{aligned}
    \label{eq:partial_dot_x}
    \frac{d}{dt} \frac{\partial L}{\partial \dot{x}^p_{\vec{i}}} &= \frac{d}{dt} \left[ \frac{M}{4} \sum_{b=0}^{d-1} \left( \dot{x}^p_{\vec{i} + \hat{e}^b} + \dot{x}^p_{\vec{i}} \right) - \frac{M}{12} \sum_{b=0}^{d-1} \left( \dot{x}^p_{\vec{i} + \hat{e}^b} - \dot{x}^p_{\vec{i}} \right) + \cdots \right. \\
    &\hspace{2cm} \cdots \left. \frac{M}{4} \sum_{b=0}^{d-1} \left( \dot{x}^p_{\vec{i}}+ \dot{x}^p_{\vec{i} - \hat{e}^b} \right) - \frac{M}{12} \sum_{b=0}^{d-1} \left( \dot{x}^p_{\vec{i}} - \dot{x}^p_{\vec{i} - \hat{e}^b} \right) \right]. 
    % &=\frac{M}{4} \frac{d}{dt} \left[ \sum_{b=0}^{d-1} \left( \dot{x}^p_{\vec{i} + \hat{e}^b} + 2\dot{x}^p_{\vec{i}} + \dot{x}^p_{\vec{i} - \hat{e}^b} \right) - \frac{1}{3} \sum_{b=0}^{d-1} \left( \dot{x}^p_{\vec{i} + \hat{e}^b} - \dot{x}^p_{\vec{i} - \hat{e}^b} \right) \right] \\
\end{aligned}
\end{equation}
If $\vec{i}+\hat{e}^b \in \mathbb{Z}_{\vec{N}}$ and $\vec{i}-\hat{e}^b \in \mathbb{Z}_{\vec{N}}$ for all $b\in\{0,\dots, d-1\}$, then
\begin{equation}
    \begin{aligned}
    \frac{d}{dt} \frac{\partial L}{\partial \dot{x}^p_{\vec{i}}} &=\frac{M}{4} \left[ \sum_{b=0}^{d-1} \left( \ddot{x}^p_{\vec{i} + \hat{e}^b} + 2\ddot{x}^p_{\vec{i}} + \ddot{x}^p_{\vec{i} - \hat{e}^b} \right) - \frac{1}{3} \sum_{b=0}^{d-1} \left( \ddot{x}^p_{\vec{i} + \hat{e}^b} - \ddot{x}^p_{\vec{i} - \hat{e}^b} \right) \right] \\
    &=\frac{M}{2} \left[  \ddot{x}^p_{\vec{i}} d + \frac{1}{3} \sum_{b=0}^{d-1} \left(  \ddot{x}^p_{\vec{i} + \hat{e}^b} + 2 \ddot{x}^p_{\vec{i} - \hat{e}^b} \right) \right].
    % &\hspace{3cm} \cdots  \left. \frac{M}{4} \sum_{\substack{
    %     \vec{j} \in \partial_{\vec{i}} \\ \vec{j}<\vec{i}}} \left(x_{\vec{i}}^p + x_{\vec{j}}^p \right) + \frac{M}{12} \sum_{\substack{
    %         \vec{j} \in \partial_{\vec{i}} \\ \vec{j}<\vec{i}}} \left(x_{\vec{i}}^p - x_{\vec{j}}^p \right) \right] \\
\end{aligned}
\end{equation}
Therefore, for all $p \in [m]$ and non-boundary $\vec{i} \in \mathbb{Z}_{\vec{N}}$,
\begin{equation}
\begin{aligned}
    \frac{M}{2} \left[  \ddot{x}^p_{\vec{i}} d + \frac{1}{3} \sum_{b=0}^{d-1} \left(  \ddot{x}^p_{\vec{i} + \hat{e}^b} + 2 \ddot{x}^p_{\vec{i} - \hat{e}^b} \right) \right] = k \sum_{j=0}^{N-1} \left( \hat{y}^p(\mathbf{u}_j) - y^p_{j} \right) \sum_{\substack{
        \vec{v} \in \mathbb{Z}_{\vec{N}} \\ \vec{t} \in \{0,1\}^d}} \mathbf{1}_{\left \{\mathbf{u}_j \in \Omega_{\vec{v}} \right \}} \  \cdots \\
        \cdots \delta_{\vec{i}, \vec{v}+\vec{t}}\prod_{b=0}^{d-1} \left[ (1-\uplambda^b_{\vec{v}}(\mathbf{u}_j)) \delta_{t^b,0} + \uplambda^b_{\vec{v}}(\mathbf{u}_j) \delta_{t^b,1} \right].
\end{aligned}
\end{equation}
For boundary $\vec{i} \in \mathbb{Z}_{\vec{N}}$, meaning one or more coordinates of $\vec{i}$ are $0$ or $N_j$, the EOM are similar but with fewer terms (e.g. for $\vec{i} = \vec{0}$, the terms in the RHS of \cref{eq:partial_dot_x} containing $\ddot{x}^p_{\vec{i} - \hat{e}^b}$ are absent). If $M \neq 0$, we can solve for $\ddot{x}^p_{\vec{i}}$,
\begin{equation}
\begin{aligned}
    \ddot{x}^p_{\vec{i}} &=  \frac{2k}{Md} \sum_{j=0}^{N-1} \left( \hat{y}^p(\mathbf{u}_j) - y^p_{j} \right) \sum_{\substack{
        \vec{v} \in \mathbb{Z}_{\vec{N}} \\ \vec{t} \in \{0,1\}^d}} \mathbf{1}_{\left \{\mathbf{u}_j \in \Omega_{\vec{v}} \right \}} \delta_{\vec{i}, \vec{v}+\vec{t}} \  \cdots  \\
        &\hspace{1cm} \cdots \prod_{b=0}^{d-1} \left[ (1-\uplambda^b_{\vec{v}}(\mathbf{u}_j)) \delta_{t^b,0} + \uplambda^b_{\vec{v}}(\mathbf{u}_j) \delta_{t^b,1} \right] - \frac{1}{3d} \sum_{b=0}^{d-1} \left(  \ddot{x}^p_{\vec{i} + \hat{e}^b} + 2 \ddot{x}^p_{\vec{i} - \hat{e}^b} \right).
\end{aligned}
\end{equation}
To add dissipation to the system, we can add a drift and diffusion term to the EOMs of the system, resulting in a linear Langevin equation.

\section{Entropy production}
\label{sec:entropyProd}

Dissipative systems such as the one described in \cref{sec:math} can be studied using stochastic thermodynamics. Irreversible processes in these systems are characterized by their change in entropy, which can be used as a proxy for the irreversibility of the system. In general, a system that changes its state over time, will have a change in entropy given by
\begin{equation*}
    \frac{dS}{dt} = \Pi(t) - \Phi(t),
\end{equation*}
where $\Pi(t)$ is the entropy production rate and $\Phi(t)$ is entropy flux out of the system. Linear Langevin systems have exact solutions for $\Pi(t)$ and $\Phi(t)$ \cite{landi2013entropy}. Namely, for a Linear Langevin system 
\begin{equation*}
 \frac{dx(t)}{dt} = Ax(t) + b + B \dot{\xi}(t),
\end{equation*}
with covariance matrix $\Theta(t) = \langle x(t) x(t)^T \rangle - \langle x(t) \rangle \langle x(t) \rangle^T$ and diffusion tensor $D=\frac{BB^T}{2}$, the entropy production rate and entropy flux are given by
\begin{align*}
    \Pi(t) &= \Tr(D \Theta^{-1}-A^{\text{q}}) + \Tr({A^{\text{q}}}^T D^{-1} A^{\text{q}} \Theta - A^{\text{q}}) + (A^{\text{q}} \langle x(t) \rangle - b^{\text{q}})^T D^{-1} (A^{\text{q}} \langle x(t) \rangle - b^{\text{q}}), \\
    \Phi(t) &= \Tr({A^{\text{q}}}^T D^{-1} A^{\text{q}} \Theta - A^{\text{q}}) + (A^{\text{q}} \langle x(t) \rangle - b^{\text{q}})^T D^{-1} (A^{\text{q}} \langle x(t) \rangle - b^{\text{q}}),
\end{align*}
where $A^{\text{q}}$ and $b^{\text{q}}$ are the components of $A$ and $b$ respectively that are invariant under time reversal $t \mapsto -t$. We can decompose functions $f$ of the dynamical system in $f(t)=f^q (t) + f^p (t)$ where $f^q(-t)=f^q(t)$ and $f^p(-t)=-f^p(t)$. In the springs and sticks system of \cref{sec:math}, $b^q=0$ and $A^q=\operatorname{Diag}(0, -\gamma \mathbf{I})$.

\begin{figure}
    \centering
    \includegraphics[width=\textwidth]{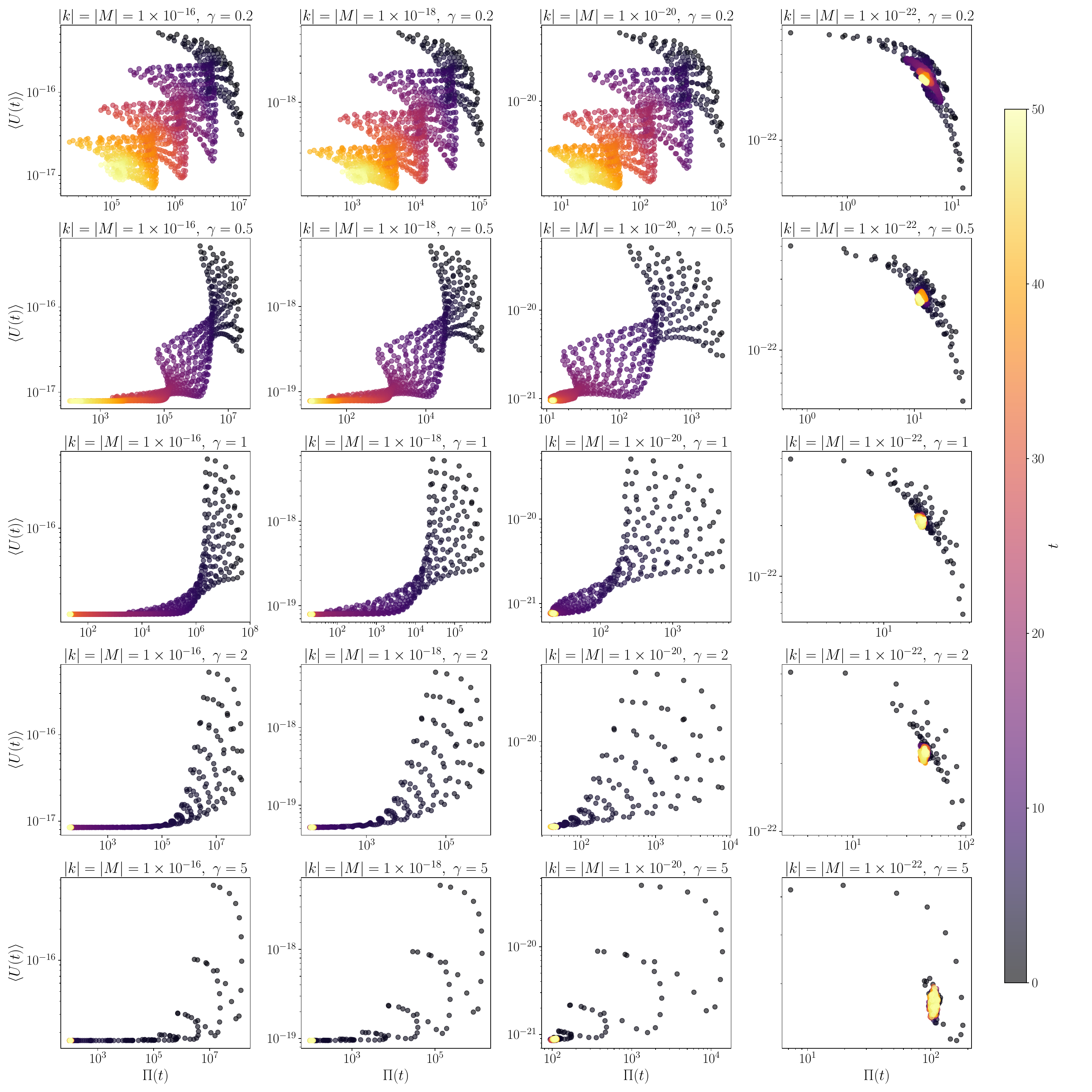}
    \caption{Entropy production rate and potential energy of the system as a function of time for different values of $\gamma$, $k$ (and $M$).}
    \label{fig:entropy_production}
\end{figure}

Using the aforementioned decomposition, we calculate the entropy production rate for the springs and sticks system composed of one stick and ten springs for data points $(x, f(x))$ where $x \sim \mathcal{U}([0,1])$ and $f(x)= 0 + \epsilon\mathcal{N}(0, 1)$, with $\epsilon=0.1$, as shown in \cref{fig:entropy_production}. We observe that the entropy production rate decreases as the potential energy of the system decreases.

% \begin{figure}
%     \centering
%     \includegraphics[width=\textwidth]{figs/entropy_flux_results.pdf}
%     \caption{Entropy flux and potential energy of the system as a function of time for different values of $\gamma$, $k$ and $M$.}
%     \label{fig:entropy_flux}
% \end{figure}

% \section{Relaxation times}

\end{document}